\pdfoutput=1

\documentclass[11pt]{article}

\usepackage[final]{acl}

\usepackage{times}
\usepackage{latexsym}
\usepackage{tabularray}
\usepackage{float}
\usepackage{tabularx,booktabs}
\UseTblrLibrary{siunitx}
\usepackage{graphicx} 

\usepackage{float}
\usepackage{stfloats}

\usepackage[T1]{fontenc}

\usepackage[utf8]{inputenc} 
\usepackage[T1]{fontenc}    
\usepackage{hyperref}       
\usepackage{url}            
\usepackage{booktabs}       
\usepackage{amsfonts}       
\usepackage{nicefrac}       
\usepackage{microtype}      
\usepackage{xcolor}         
\usepackage{color}
\usepackage{babel,blindtext}
\usepackage{subcaption}
\usepackage{algorithm} 
\usepackage{latexsym}
\usepackage{caption}
\usepackage{algpseudocode} 
\usepackage{makecell}
\usepackage{graphicx} 
\usepackage{multirow} 
\usepackage{hhline} 
\usepackage{amsmath} 
\usepackage{array,booktabs}

\definecolor{myblue}{RGB}{92,187,211}
\definecolor{mygreen}{RGB}{69,141,88}
\definecolor{myorange}{RGB}{219,110,59}
\definecolor{mypurple}{RGB}{113,76,165}

\definecolor{Gray}{gray}{0.9}

\usepackage{microtype}

\usepackage{inconsolata}

\title{MORL-Prompt: An Empirical Analysis of Multi-Objective Reinforcement \\ Learning for Discrete Prompt Optimization}

\author{
  Yasaman Jafari $\quad$ Dheeraj Mekala $\quad$ Rose Yu $\quad$  Taylor Berg-Kirkpatrick  \\ \\
  University of California San Diego\\
  \small \texttt{\{yajafari, dmekala, roseyu, tberg\}@ucsd.edu}
}

\begin{document}
\maketitle

\begin{abstract}

    RL-based techniques can be employed to search for prompts that, when fed into a target language model, maximize a set of user-specified reward functions. 
    However, in many target applications, the natural reward functions are in tension with one another -- for example, content preservation vs. style matching in style transfer tasks. 
    Current techniques focus on maximizing the average of reward functions, which does not necessarily lead to prompts that achieve\textit{ balance across rewards} -- an issue that has been well-studied in the multi-objective and robust optimization literature. 
    In this paper we conduct an empirical comparison of several existing multi-objective optimization techniques, adapted to this new setting: RL-based discrete prompt optimization. We compare two methods optimizing the volume of the Pareto reward surface, and one method that chooses an update direction that benefits all rewards simultaneously.
     We evaluate performance on two NLP tasks: style transfer and machine translation, each using three competing reward functions.
    Our experiments demonstrate that multi-objective methods that directly optimize the volume of the Pareto reward surface perform better and achieve a better balance of all rewards than those that attempt to find monotonic update directions.
    

\end{abstract}

\section{Introduction}




Discrete prompt tuning involves refining a text prompt for a language model (LM) to maximize a set of user-specified objectives on the LM's output~\cite{shin-etal-2020-autoprompt, Schick2020ExploitingCF, wen2023hard}. Successful techniques for prompt tuning allow users to control and adapt powerful LLMs to new tasks without the trial-and-error of manual prompt design.  While RL-based techniques have been shown to be effective at finding prompts that optimize an average of rewards~\cite{deng-etal-2022-rlprompt}, in many target applications, there is a tension between the natural reward functions. 

\begin{figure}
    \centering
    \includegraphics[width=0.47\textwidth]{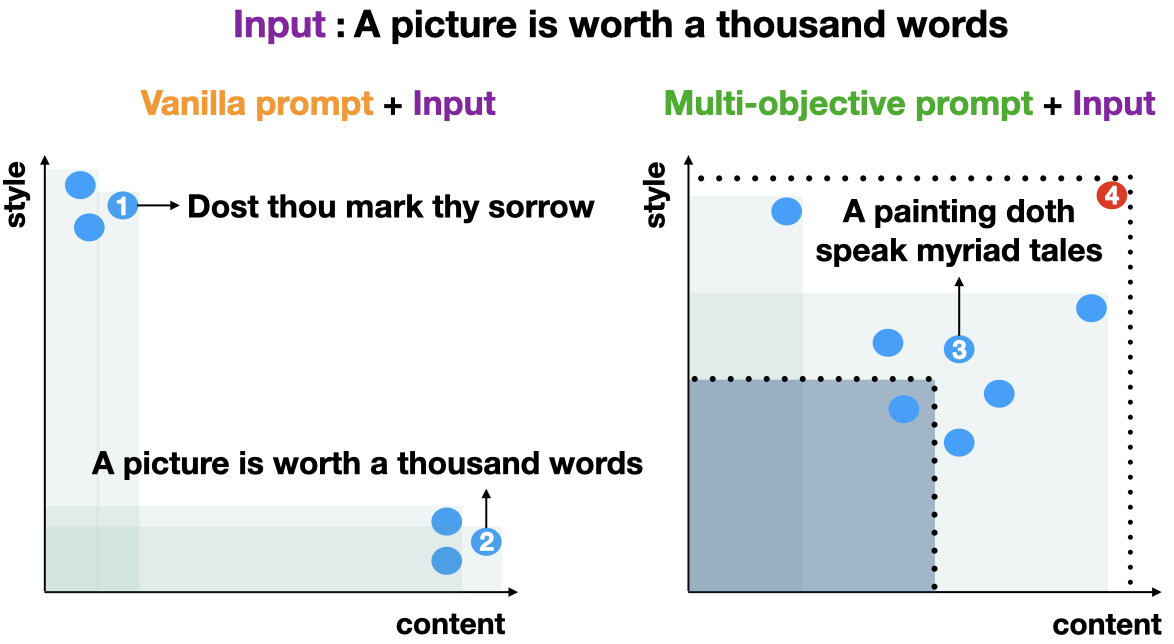}
    \caption{A modern to Shakespearean text style transfer setting where each dot represents an output sentence sampled from an LM conditioned on either a prompt trained with average reward (left) or a prompt trained using multi-objective optimziation techniques (right). The output sample 1 only optimizes for style match, while output sample 2 only addresses content preservation. Sample 3, on the other hand, balances both objectives at the same time. The shaded regions indicate measures of volume of the Pareto reward surface.}
    \label{fig:example}
\end{figure}

For example, as depicted in Figure~\ref{fig:example}, many style transfer tasks need to preserve content while simultaneously maximizing transfer into the target style -- two objectives that are directly at odds with one another. Thus, current techniques result in a phenomenon we will refer to as \textit{objective collapse}: focusing on maximizing the average of reward functions (also called \textit{scalarization}) can lead to prompts that disproportionately maximize a subset of objectives at the expense of others. For instance, in Figure \ref{fig:example}, the prompt on the left side tends to produce LM outputs (represented by blue dots) that prioritize one objective over the other. Conversely, the prompt on the right side produces samples that achieve reasonable performance across all objectives simultaneously. However, in both cases the \textit{average reward} is nearly equivalent. 



The problem of reward balancing has been well-studied in other domains---for example, the multi-objective and robust optimization literature proposes several approaches that offer advantages over scalarization. However, these techniques have not been applied to the RL-based discrete prompt optimization setting that is most relevant in NLP. Thus, in this paper we conduct an empirical comparison of several existing techniques for multi-objective optimization that we adapt to discrete prompt optimization, where we aim to evaluate their effectiveness in achieving a more useful balance of rewards in downstream prompt-driven NLP tasks. 
The first two approaches we compare maximize the volume of the Pareto reward surface, while the third method chooses a gradient update direction that is beneficial for all rewards simultaneously.


\begin{figure*}
    \centering
    \includegraphics[width=0.6\textwidth]{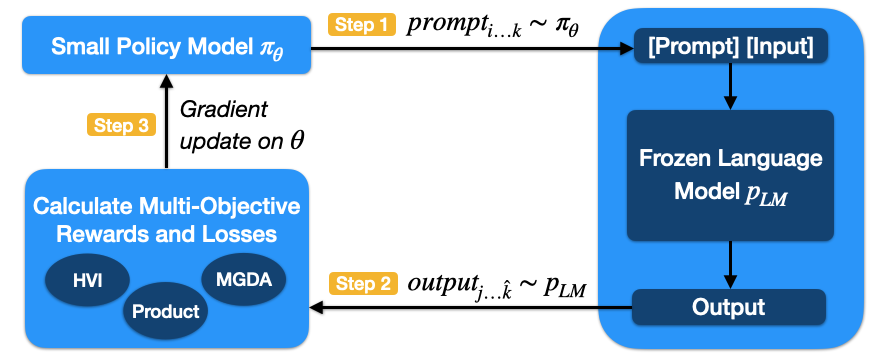}
    \caption{In all the settings, we have a parameter-efficient policy model, responsible for generating the task-specific prompts, where all the parameters of the model except for an MLP module are frozen. Another frozen language model is used to generate output sentences, given an input and a prompt from the policy model. All the output sentences are then evaluated with respect to each objective, and multi-objective losses are calculated. Finally, a gradient update on the MLP parameters is performed. }
    \label{fig:model-outline}
\end{figure*}

More specifically, the first method in our study computes the hypervolume indicator (HVI)~\cite{knowles2004bounded} for a set of samples drawn from a given prompt, and treats this measure as the final reward in RL. 
Intuitively, HVI measures the area under the Pareto frontier of the outputs sampled from the current prompt (shown by the shaded regions in Figure~\ref{fig:example}). 
Samples that achieve a better balance of reward elevate the Pareto frontier and increase HVI.
However, this method has a potential drawback: if an outlier sample (e.g., represented by the red dot labeled with a four in Figure~\ref{fig:example}) achieves high values across all rewards, the HVI can be disproportionately high (represented by the outer rectangular region in Figure~\ref{fig:example}, which dominates the shaded areas). 
This \textit{dominant outlier effect} may diminish the stability of HVI optimization in an RL setting, as it becomes very sensitive to outliers. 

Therefore, we also investigate using a simpler method for maximizing the volume in the second approach, called the expected product of rewards. Here, we approximate the expected volume by simply computing the average \textit{product of rewards} (tentatively depicted by the dark rectangular region in Figure~\ref{fig:example}).

The third approach takes a different strategy based on \textit{steepest gradient descent} ~\cite{fliege2000steepest}. 
We approximate the gradient of the expectation of each individual reward separately and then search for an update direction to make monotonic progress in every reward simultaneously. 


To understand the effectiveness of these approaches in the discrete prompt optimization setting, we conduct experiments on two text generation tasks: text style transfer and machine translation using three competing reward functions for each task. 
Our findings indicate that volume-based methods are most effective in this setting, achieving substantial gains in balancing the competing rewards, compared to the baseline methods. 
While RL-based steepest descent also improves balance, it does not perform as robustly as the volume-based methods.

\begin{figure*}[t]
    \begin{algorithmic}[1]
        \Statex \hrule
        \vspace{-0.1cm}
        \Statex \hspace{-0.7cm} \textbf{Algorithm 1:} Volume-based policy update for one input sentence.
        \vspace{0.1cm}
        \Statex \hrule
        \State Input: Input sentence $x$, policy $\pi_{\theta}$, reward models $r_{1\dots m}$, external frozen LM  
                \State $\{z_{1\dots k}\} \sim \pi_\theta(x)$ \Comment{Sample k prompts from the policy}
            \For{i = 1$\dots$k}:
                 \State $\{y_{1\dots \hat{k}}\} \sim p_{LM}(y|x,z_{i})$ \Comment{Sample $\hat{k}$ output sentences from a desired LM}
            \EndFor
            \For{i = $1\dots k\cdot \hat{k}$ \do}:
            \State calculate $r_{1\dots m} (y_{i}, x)$ \Comment{Calculate $r_{1\dots m}$ for each output sentence y and input x}
            \EndFor
            \State Calculate $r_{prod}$ (or $r_{hvi}$)
            \Comment{Calculate expected product of rewards (or hypervolume)}
            \State Calculate ${\mathcal{L}}$ using $r_{prod}$ (or $r_{hvi}$) \Comment{Use SQL loss \cite{guo2022efficient}}
            \State $\theta=\theta-\eta \nabla_{\theta} {\mathcal{L}}\left(\theta\right)$ \Comment{Gradient descent on policy parameters}
            \vspace{0.2cm}
            \Statex \hrule
    \end{algorithmic} 
    \caption{In this algorithm, k prompts are sampled from the policy model and used alongside an input sentence to generate $\hat{k}$ output samples from an external frozen LM. The desired objective values for each of the sentences are calculated and combined into a single reward value by computing their expected product of rewards or hypervolume indicator, based on the desired approach. Then, the loss is computed based on this reward and used for a gradient update on the policy LM's parameters. For details on the SQL loss and parameter updates, we refer the interested reader to \cite{guo2022efficient}.}
    \label{fig:vol}
\end{figure*}

\section{Problem Statement}
In this paper, we specifically focus on optimizing discrete prompts, as they offer the advantages of interpretability and reusability in contrast to continuous or ``soft'' prompts. We acknowledge that the issue of balancing multiple conflicting objectives is a well-established area of research within the multi-objective and robust optimization literature. However, adapting these techniques to the domain of discrete prompt optimization for language models comes with challenges due to having sources of discontinuity and discreteness. First, the text tokens for the prompt are discrete, and second, since marginalizing over all possible output samples is intractable, we need to approximate the expected gradient of the loss with respect to the sampled sentences. 

We train a small, parameter-efficient policy network in order to generate task-specific prompts that can later be used alongside an input sentence to be fed into any other language model, as depicted in Figure \ref{fig:model-outline}. We particularly put an emphasis on optimizing prompts to achieve a balance across multiple reward functions.
Given $m$ multiple objectives and their corresponding reward functions $\{r_1, r_2, \ldots r_m\}$, we perform discrete prompt optimization for controlled text generation.
We refer to the prompt as $z$, the input text as $x$, and the text generated by the LM as $y$. 
We aim to generate a prompt that is added to the beginning of the input and causes the LM to generate output text compliant with the objectives. 

\subsection{Optimization problem}
We formulate discrete prompt optimization as an RL problem, where we train a multi-layer perceptron (MLP) module over a frozen language model as our policy network. A frozen LM head is used after the MLP module to generate the prompts.

The RL-based approach to discrete prompt optimization tries to optimize an intractable objective through stochastic approximation, and a common way of incorporating multiple objectives is to use their sum (scalarization). In Equation \ref{eqn:problem-statement}, we show the intractable objective for scalarization that RL-based methods attempt to optimize. 

\begin{equation}
\label{eqn:problem-statement}
\max_\theta \mathbb{E}_{z \sim \pi_\theta}\left[\mathbb{E}_{y\sim p_{LM}(y \mid x, z)}\left[\sum_{i=1}^m r_i\left(y, x\right)\right]\right]
\end{equation}

Note that Equation \ref{eqn:problem-statement} involves true expectations, which are intractable to compute; therefore, we approximate the expectations by utilizing samples from the policy and the frozen language model. At each step, given a text input $x$, we sample $k$ prompts $\{z_1, z_2, \ldots z_k\}$ from the policy $\pi_{\theta}$, where $\theta$ represents the policy parameters. 
Subsequently, we utilize another frozen language model $p_{LM}$ to generate $\hat{k}$ output sentences for each pair of input $x$ and prompt $z_i$. Then, we assess the quality of these outputs using the reward function $r_i$ corresponding to each objective\footnote{For simplicity, we assume the reward value is solely dependent on the generated text $y$ and the input text $x$. It can be easily expanded to include prompt $z$ or the reference text, if necessary.}.

We will explore RL-based approaches that go beyond simply summing rewards, which may offer better ways to balance multiple objectives.

\section{Methodology}
In this section, we describe the adapted optimization methods for generating discrete prompts that, when fed into an LM along with the input text, produce outputs that maximize a set of competing reward functions.
We compare two optimization methods that maximize the volume coverage of rewards and one method that finds the gradient update direction which optimizes all rewards simultaneously.

We adopt the soft Q-learning (SQL) reinforcement learning framework introduced by \cite{guo2022efficient}, which has demonstrated effectiveness in discrete prompt optimization with single reward functions or scalarization \cite{deng-etal-2022-rlprompt}. In line with \cite{deng-etal-2022-rlprompt}, we utilize only the on-policy component of SQL. For clarity and simplicity, we omit certain details of the SQL updates in the pseudo-codes (Figures \ref{fig:vol} and \ref{fig:mgda}), focusing instead on the novel multi-objective components. For a comprehensive discussion of SQL, we refer the reader to \cite{guo2022efficient}.

\subsection{RL-based Volume Improvement}
In this section, we investigate two approaches that aim to improve the volume coverage of rewards.
\subsubsection{Hyper-volume indicator}
The hypervolume indicator ~\cite{zitzler1998multiobj, knowles2004bounded} is defined for a point set $S \subset \mathbb{R}^d$ and a reference point $p_{ref} \in \mathbb{R}^d$. 
The hypervolume indicator $H$ quantifies the region dominated by $S$ and bounded by $p_{ref}$. 
S denotes the set of points/solutions that we are examining. 
Mathematically, hyper-volume indicator is defined as:

\[
H(\mathrm{~S})\hspace{-0.1cm}=\hspace{-0.1cm}\Lambda\hspace{-0.1cm}\left(\left\{q \in \mathbb{R}^d \hspace{-0.05cm}\mid\hspace{-0.05cm} \exists p \in \mathrm{S}\hspace{-0.05cm}:\hspace{-0.05cm} q\hspace{-0.05cm} \leq\hspace{-0.05cm} p \text{ and } p_{ref}\hspace{-0.05cm} \leq \hspace{-0.05cm} q\right\}\right)
\]

where the notation q $\leq$ p means that each component of the vector q is less than or equal to each of the corresponding components of the vector p, and $\Lambda(\cdot)$ shows the Lebesgue measure for the sub-space. In other words, $\Lambda(\cdot)$ measures the size of the hypervolume covered by a set of solutions in the objective space. This hypervolume is always measured with respect to a reference point, which we consider to be a zero vector in all our experiments.

In our setting, each point in $S$ is a sampled sentence. 
For example, in the style-transfer task, if we have 2 objective values of style-match: 0.6 and content-match: 0.3 for a sentence, this point can be denoted as (0.6, 0.3), and the reference point would be set to (0, 0). 
We consider the hypervolume indicator of the reward functions as the ultimate reward signal for training the policy network in the first approach.



\begin{figure*}[t]
    \begin{algorithmic}[1]
        \Statex \hrule
        \vspace{-0.1cm}
        \Statex \hspace{-0.7cm} \textbf{Algorithm 2:} MGDA-based policy update for one input sentence.
        \vspace{0.1cm}
        \Statex \hrule
        \State Input: Input sentence $x$, policy $\pi_{\theta}$, reward models $r_{1\dots m}$, external frozen LM                  \State $\{z_{1\dots k}\} \sim \pi_\theta(x)$ \Comment{Sample k prompts from the policy}
            \For{i = 1$\dots$k}:
                 \State $\{y_{1\dots \hat{k}}\} \sim p_{LM}(y|x,z_{i})$ \Comment{Sample $\hat{k}$ output sentences from a desired LM}
            \EndFor
            \For{i = $1\dots k\cdot \hat{k}$ \do}:
            \State calculate $r_{1\dots m} (y_{i}, x)$ \Comment{Calculate $r_{1\dots m}$ for each output sentence y and input x}
            \EndFor
            \For{i = 1$\dots$m}:
                \State Calculate ${\mathcal{L}}_m$ using $r_m$ \Comment{Use SQL loss \cite{guo2022efficient}}
            \EndFor
            \State $\lambda_1, \ldots, \lambda_m=$ FrankWolfeSolver$(\nabla_{\theta} {\mathcal{L}}_i\left(\theta\right))$ \Comment{Find the direction using [\ref{eqn:dual}}]
            \State $\theta=\theta-\eta \sum_{i=1}^m \lambda_i \nabla_{\theta} {\mathcal{L}}_i\left(\theta\right)$ \Comment{Gradient descent on policy parameters}
            \vspace{0.2cm}
            \Statex \hrule
    \end{algorithmic} 
    \caption{In this algorithm, k prompts are sampled from the policy model and used alongside an input sentence to generate $\hat{k}$ output samples from an external frozen LM. The desired objective values for each of the sentences are calculated and used to generate the corresponding losses. Then, a direction to improve all the losses at the same time is found, and a gradient update on the policy model's parameters is performed.}
    \label{fig:mgda}
\end{figure*}

\subsubsection{Expected product of rewards}
In this method, we consider the expected product of objective functions as the reward signal for training the policy network.
We obtain $\hat{k}$ samples as output per prompt and for each sentence, we compute all $m$ reward values, and calculate the product of rewards. 
We utilize the expected product of rewards across all $\hat{k}$ samples as the final reward signal for policy updates.

The main advantage of this reward compared to the HVI reward is that the effect of the outliers will be more controlled by using the expected value of objectives within a sampled set of sentences. 

The pseudo-code for the volume-based approaches is provided in Figure \ref{fig:vol}, where at each update step, we sample prompts from the policy model and generate output sentences from a desired language model. We then calculate the reward values for each of the objectives separately and use them to compute the dominated hypervolume or the expected product of rewards and use it to calculate the loss. Then, we update the policy model using gradient descent.

\subsection{Multiple Gradient Descent Algorithm with RL}
We describe the multiple gradient descent algorithm (MGDA), which finds the gradient update direction that maximizes all the rewards.
This method follows the approach of steepest descent for multi-criteria optimization ~\cite{fliege2000steepest}, where the goal is to find a direction $d_t$ that improves all the objectives by the amount of $\alpha_t$, at each step $t$. Here, $L_i$ and $\theta$ represent the expected loss corresponding to objective $i$, and the parameters of the policy model, respectively.
\begin{equation}
\label{eqn:steep-gradient-descent}
\begin{aligned}
    &\left(d_{t}, \alpha_{t}\right)=\arg \min _{d \in \mathbb{R}^n, \alpha \in \mathbb{R}} \alpha+\frac{1}{2}\|d\|^2, \\
    &\text{s.t. } \nabla \mathcal{L}_i\left(\theta_t\right)^T d \leq \alpha, \quad i=1, \ldots, m .
\end{aligned}
\end{equation}

The update rule for the parameters $\theta$ at time $t$ with the step size $\eta$ is defined as:
\begin{equation}
    \theta_{t+1}=\theta_{t}-\eta d_{t}
\end{equation}

This approach has been used in continuous multi-objective settings \cite{sener2019multitask, lin2019pareto}. 
However, in our setting, since we optimize for discrete prompts, we compute stochastic gradient approximations by sampling LLM outputs and then use reinforcement learning to estimate the gradient based on the samples.
We calculate all $m$ rewards for each (prompt $z$, input $x$, generated text $y$) triplet and optimize them. 

The pseudo-code for this approach is provided in Figure \ref{fig:mgda}, where at each update step, we start by sampling prompts from the policy model and use them to generate output sentences from another language model. We then calculate the reward values for each of the objectives separately and compute their corresponding losses. Then, a direction for improving all the losses simultaneously is calculated and used for the policy update. 
More details are available in Appendix $\S$\ref{app:mgda}. 

\section{Experiments}

We now describe the empirical comparison of the RL-adapted multi-objective optimization methods that we introduced in the previous section. Our primary aim is to evaluate these techniques for discrete prompt optimization for downstream generative NLP tasks. Based on the availability of benchmarks and evaluation metrics, we focus on style transfer and machine translation tasks.
\subsection{Tasks \& Datasets}
In this section, we describe the tasks, datasets, and their corresponding competing objectives.
We evaluate on two tasks: unsupervised text style transfer and supervised machine translation. 

\begin{table*}
\centering
\scalebox{1.0}{
\begin{tabular}{c c c c c c}
\toprule
\textbf{Method} &  \textbf{Obj 1} & \textbf{Obj 2} & \textbf{Obj 3} & \textbf{Product} & \textbf{Average} \\
\midrule
\multicolumn{6}{l}{\textcolor{gray}{Text Style Transfer ($Obj_1$: Content - $Obj_2$: Style - $Obj_3$: Sentiment)}} \\
\midrule
 Average & 19.56 & \textbf{79.25} & 38.28 & 30.91 & \textbf{45.69} \\
 Product & \textbf{34.58} & 57.78 & 35.11 & \textbf{36.04} & 42.49 \\
 HVI & 25.39 & 67.91 & \textbf{38.76} & 32.44 & 44.02 \\
 MGDA & 22.37 & 66.51 & 38.11 & 31.16 & 42.33 \\
\midrule
\multicolumn{6}{l}{\textcolor{gray}{Machine Translation ($Obj_1$: Content - $Obj_2$: BLEU - $Obj_3$: Sentiment)}} \\
\midrule
 Average & 32.07 & \textbf{32.00} & 46.36 & 65.48 & 36.81 \\
 Product & \textbf{32.95} & 31.70 & 46.47 & \textbf{65.98} & \textbf{37.04} \\
 HVI & 31.18 & 30.51 & \textbf{48.69} & 63.21 & 36.79 \\
 MGDA & 31.46 & 31.85 & 46.03 & 62.87 & 36.45 \\
 \bottomrule
\end{tabular}
}
\caption{Reward values corresponding to each objective at a checkpoint where each method achieved the highest average of the product of rewards across samples. Even though the method utilizing the average of rewards achieved the highest average value for style transfer, we can observe an imbalance across various objective values. The product method, on the other hand, got the highest expected product value, reflecting a more balanced improvement. All the reported values are average objective values computed from 128 output samples. }
\label{tab:comparison}
\end{table*}

We consider hypothetical tasks such as conveying positive sentiment as a competing objective in addition to accurate style transfer or machine translation. The selection of these specific objectives and tasks is motivated by the availability of standard evaluation datasets and well-established metrics within the NLP community. For style transfer, we focus on a specific sub-task that is well-supported by available ground-truth parallel style transfer data. Specifically, we aim to transfer modern English into a Shakespearean style. This particular style transfer task has long been a mainstay benchmark for the text style transfer NLP community \cite{he2020APF, deng-etal-2022-rlprompt}. 

\paragraph{Unsupervised Text Style Transfer.} We experiment on the style transfer task ~\cite{xu2012paraphrasing, jin2022deep}, converting standard English into Shakespearean style. 
We consider three competing objectives: maintaining the original content of the input text, infusing it with Shakespearean style, and ensuring the resulting text conveys a positive sentiment.
We test on the Shakespeare dataset \cite{xu2012paraphrasing, jhamtani2017shakespearizing}, and
the objective function corresponding to content preservation is BertScore \cite{zhang2020bertscore}, for sentiment is a sentiment RoBERTa-base classifier\footnote{cardiffnlp/twitter-roberta-base-sentiment-latest}, and for style is a DistilBERT-base-uncased model fine-tuned on Shakespearean data\footnote{notaphoenix/shakespeare\_classifier\_model}.

It is noteworthy that while Shakespearean style and positive sentiment may not directly be in conflict, they are not correlated either. Shakespeare’s work includes many tragedies, such as “Hamlet,” \cite{shakespeare1703tragedy} “Macbeth,” \cite{shakespeare1710macbeth} etc., often with negative sentiments. On the other hand, Shakespeare has Comedies like “A Midsummer Night's Dream” \cite{shakespeare1734midsummer} that are more positive in terms of sentiment. Further, operationally, the way the objectives can conflict is that there may be word changes that easily improve sentiment reward (e.g., "AWESOME") that break the style reward and vice versa.

\paragraph{Supervised Machine Translation.} We experiment on German to English translation task, using the \emph{iwslt2017} data \cite{cettolo2017overview}. The objectives and the reward functions are: (1) semantic similarity between the generated translation and a reference text computed using BertScore,
(2) BLEU score \cite{papineni02bleu} between generated text and reference, and
(3) conveying a positive sentiment quantified by the same RoBERTA-base classifier used in style-transfer task.

\paragraph{Evaluation Metrics}
We evaluate each task using its corresponding objective functions, with the goal of optimizing all rewards in a balanced manner. 
To quantify this balance, we assess performance by calculating both the mean and the expected product of the individual objectives for each task.


\subsection{Training Details}

Following \cite{deng-etal-2022-rlprompt}, we consider a multi-layer perception module on top of a small frozen distilGPT-2 model~\cite{sanh2019distilbert}, alongside a frozen LM head as the policy network. We employ an MLP with 3.1 million parameters. For the text style transfer task, we use a learning rate of 5e-5, while for the translation task, we set the learning rate to 1e-4. In both cases, we utilize the Adam optimizer.
The policy network is trained for 12,000 steps. The number of training samples used for text style transfer and machine translation are 100 and 200, respectively.
At each step, we sample eight prompts for a given input from the policy network, each comprising five tokens.
Subsequently, we feed each prompt along with its corresponding input text into a separate LM to generate 128 output samples.
We use GPT-2~\cite{Radford2019LanguageMA} for text style transfer and flan-T5-small~\cite{flant5small} for machine translation tasks.

Our choice of models was informed by an assessment of their respective strengths and capabilities in specific tasks. For instance, we observed that the flan-T5-small model exhibited superior performance in machine translation tasks compared to the GPT-2 model \cite{haddow2022survey}; we followed past work in using the base models that tended to have a reasonable starting performance on the respective tasks. For instance, T5 has been repeatedly shown to be effective at translation tasks, while GPT-2 fails to produce translations reliably. Further, we wanted to demonstrate that multi-objective optimization approaches could generalize across both the encoder-decoder and decoder-only language models. Moreover, there is a specific reason why we chose a weaker model like GPT-2 as it provides a better benchmark for multi-objective optimization precisely because GPT-2 is a weaker style transfer model out of the box compared to more recent models. As a result, there is a higher burden placed on the discrete prompt itself in order to achieve good results.

Furthermore, we employ ``Efficient (soft) Q-learning'' \cite{guo2022efficient} to learn the policy network's parameters based on the reward using gradient descent.

We repeat each experiment with three distinct random seeds and report the average results. Using NVIDIA RTX A6000, each experiment takes about 20-24 hours. 

\begin{figure*}[!htb]
\minipage{0.33\textwidth}
  \includegraphics[width=1\linewidth, height=3cm]{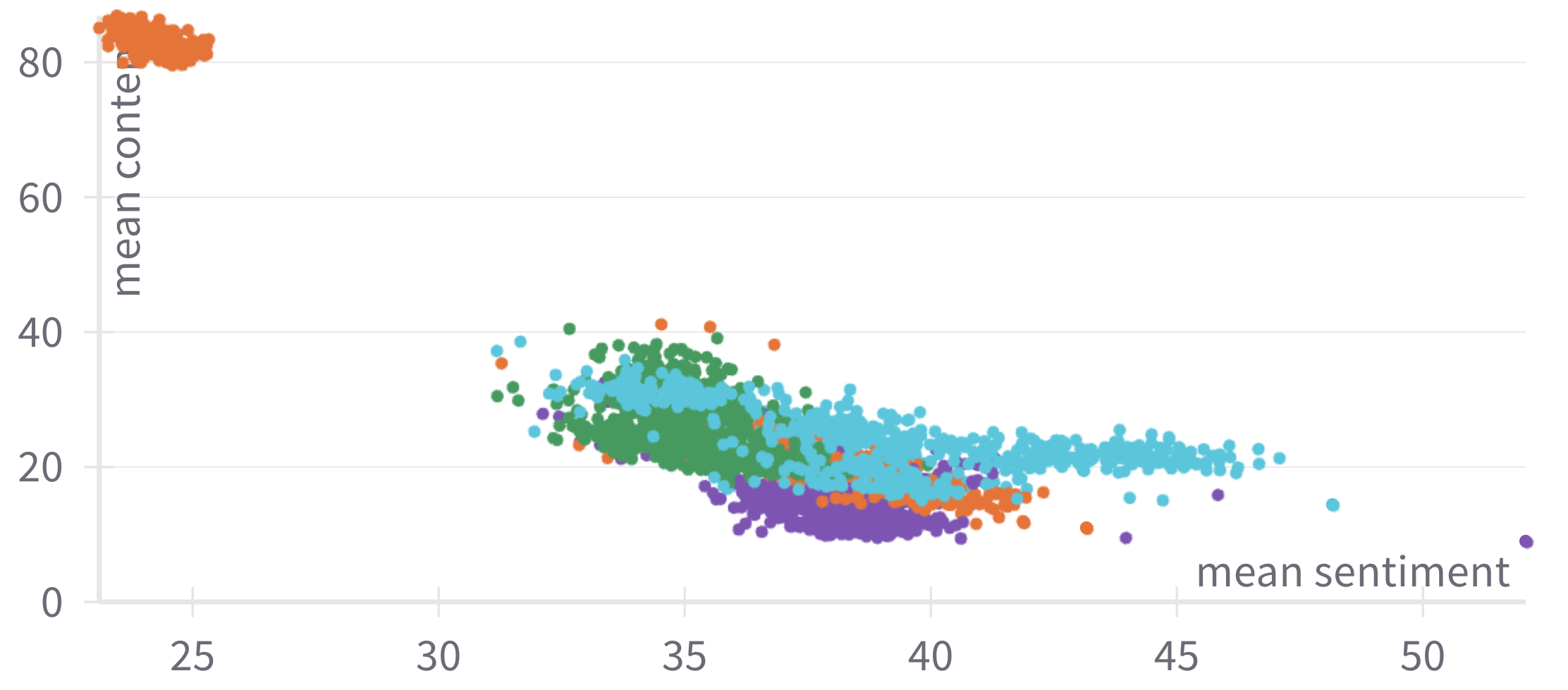}
  \caption*{Sentiment vs Content}\label{fig:tst_sentiment_content}
\endminipage\hfill
\minipage{0.33\textwidth}
  \includegraphics[width=1\linewidth, height=3cm]{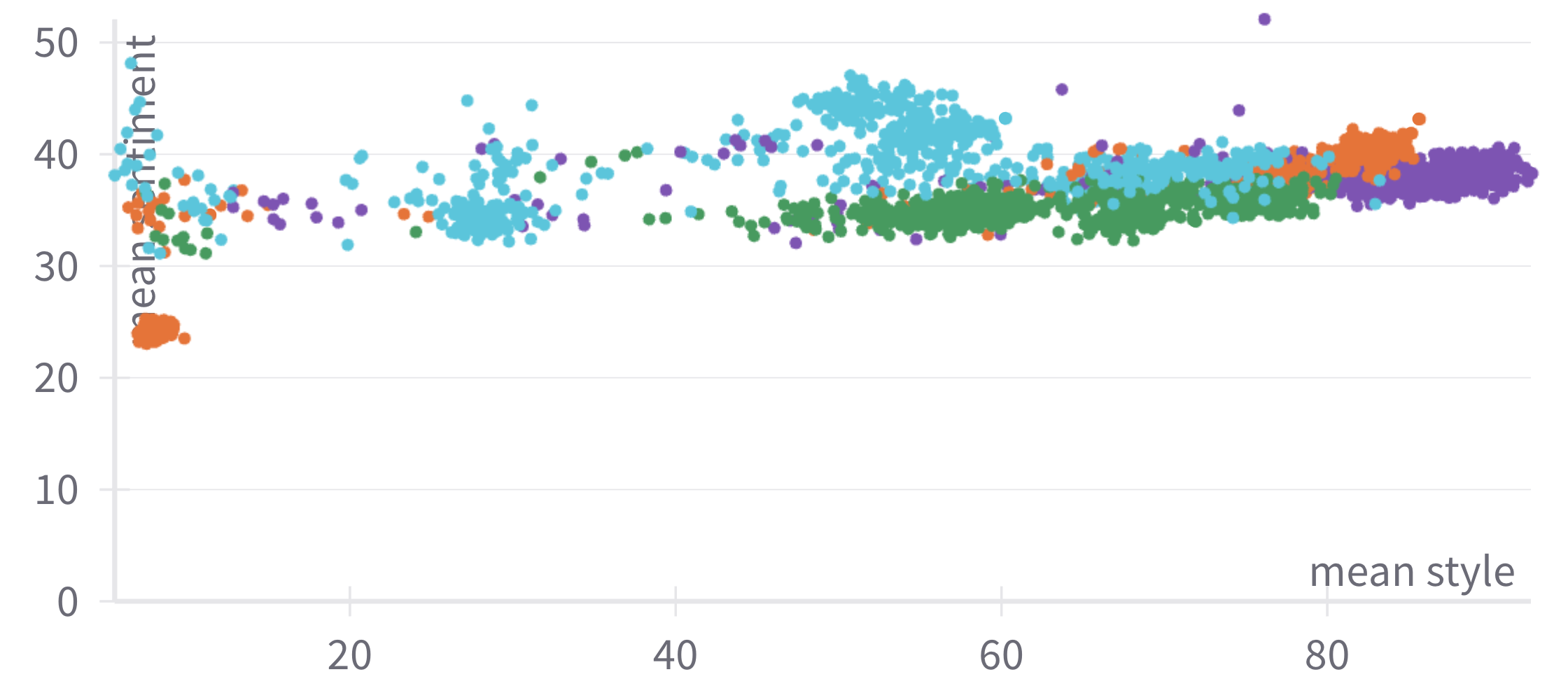}
  \caption*{Style vs Sentiment}\label{fig:tst_style_sentiment}
\endminipage\hfill
\minipage{0.33\textwidth}%
  \includegraphics[width=1\linewidth, height=3cm]{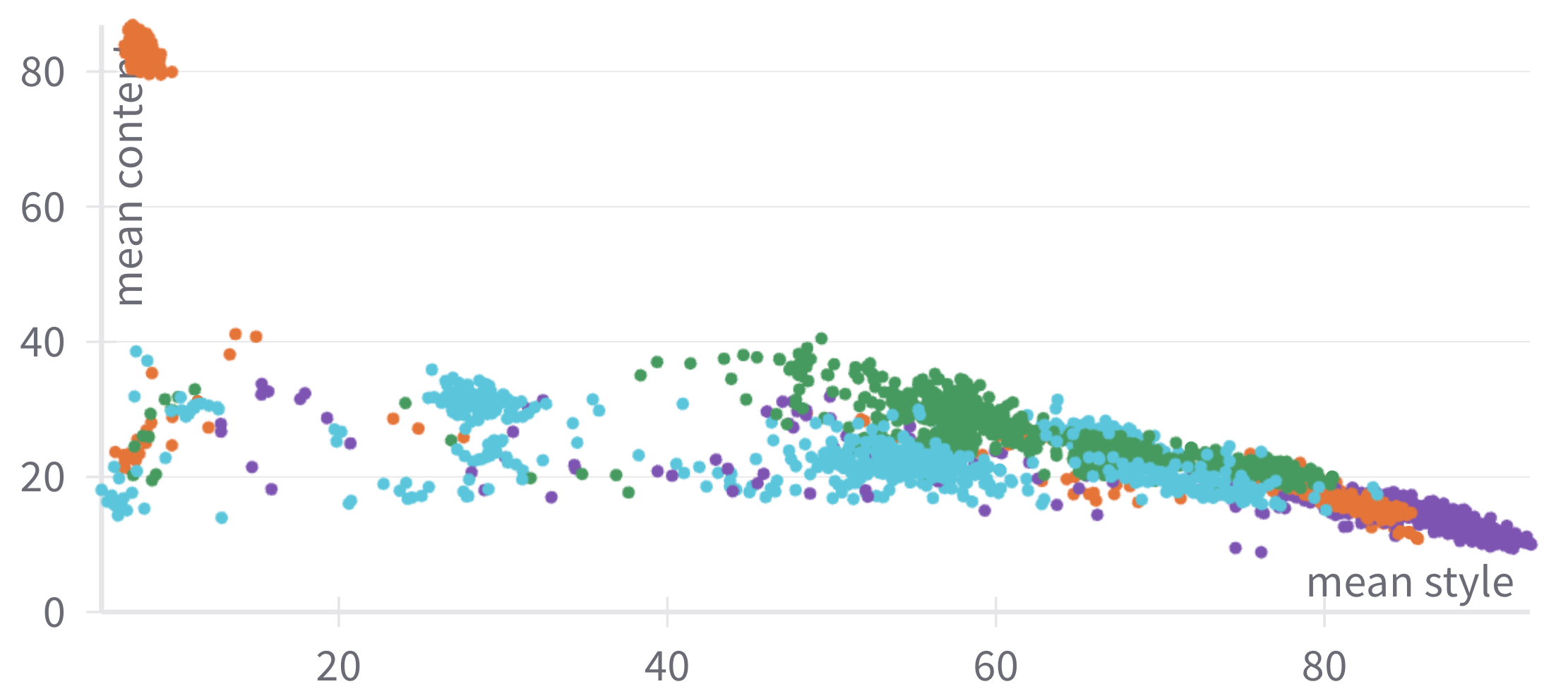}
  \caption*{Style vs Content}\label{fig:tst_style_content}
\endminipage
\caption{Text Style Transfer. From left to right, positive sentiment vs. content match, Shakespearean style vs. positive sentiment, and Shakespearean style vs. content match for different settings of \textcolor{myorange}{average reward}, \textcolor{myblue}{hyper volume indicator reward}, \textcolor{mygreen}{expected product reward}, and \textcolor{mypurple}{multiple gradient descent algorithm} are shown.}  
\label{fig:tst-pairwise}
\end{figure*}

\begin{figure*}[!htb]
\captionsetup{justification=centering}
\minipage{0.33\textwidth}
  \includegraphics[width=1\linewidth, height=3cm]{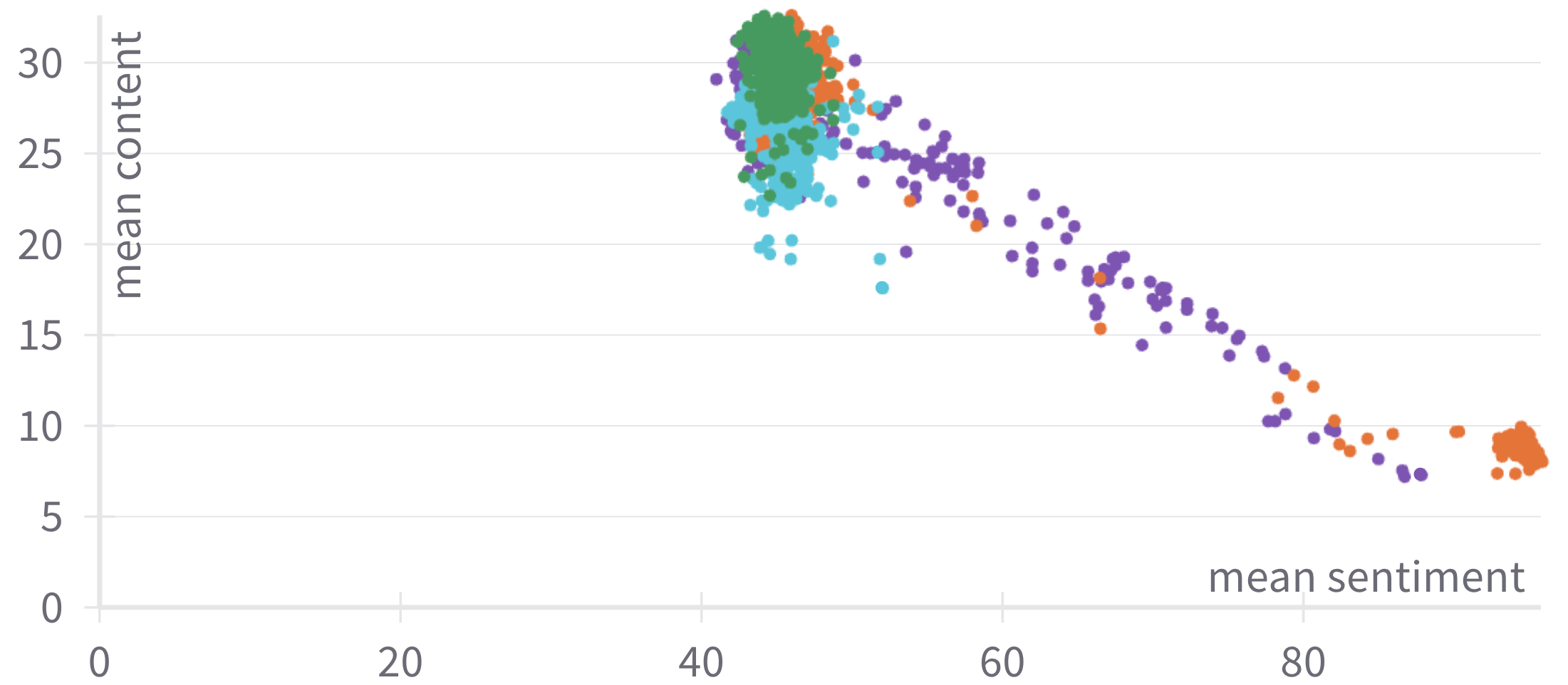}
  \caption*{Sentiment vs Content}\label{fig:mt_sentiment_content}
\endminipage\hfill
\minipage{0.33\textwidth}
  \includegraphics[width=1\linewidth, height=3cm]{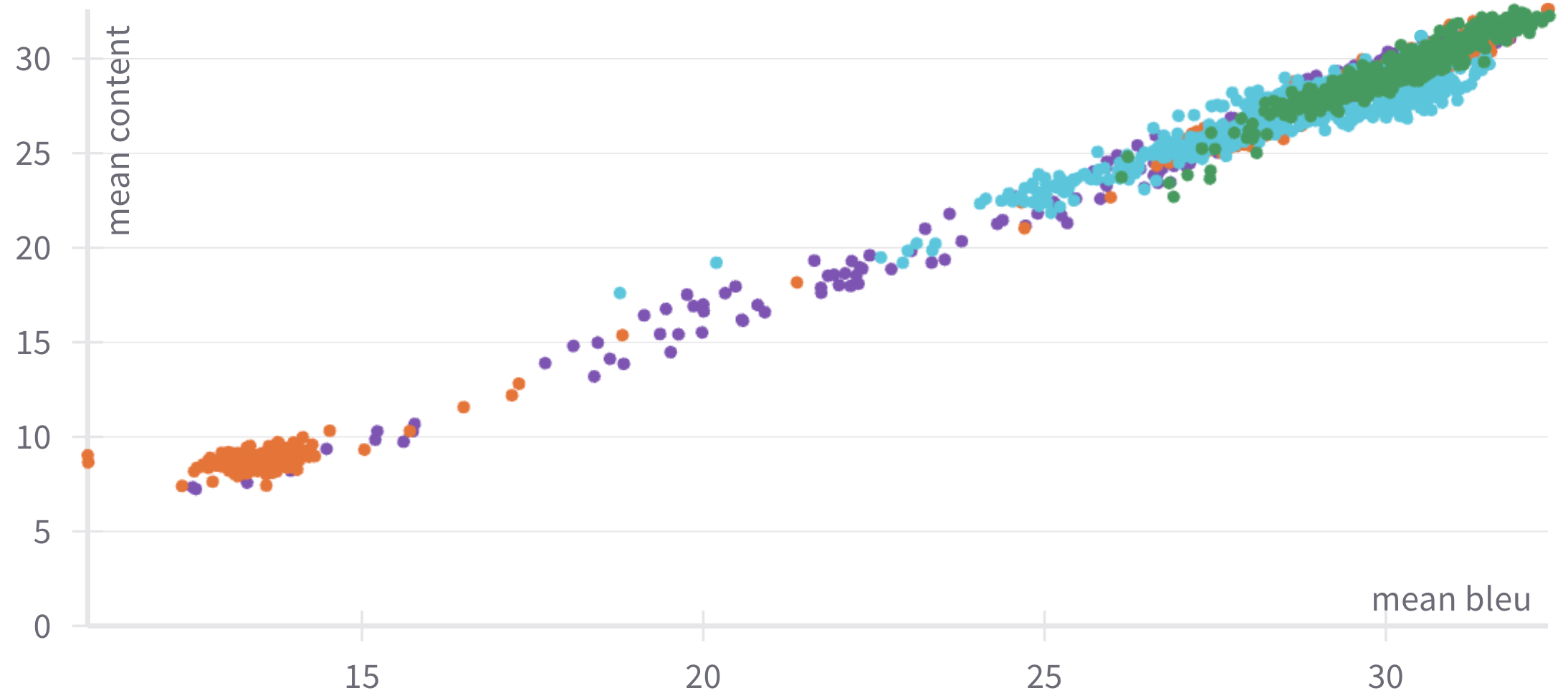}
  \caption*{BLEU vs Content}\label{fig:mt_bleu_content}
\endminipage\hfill
\minipage{0.33\textwidth}%
  \includegraphics[width=1\linewidth, height=3cm]{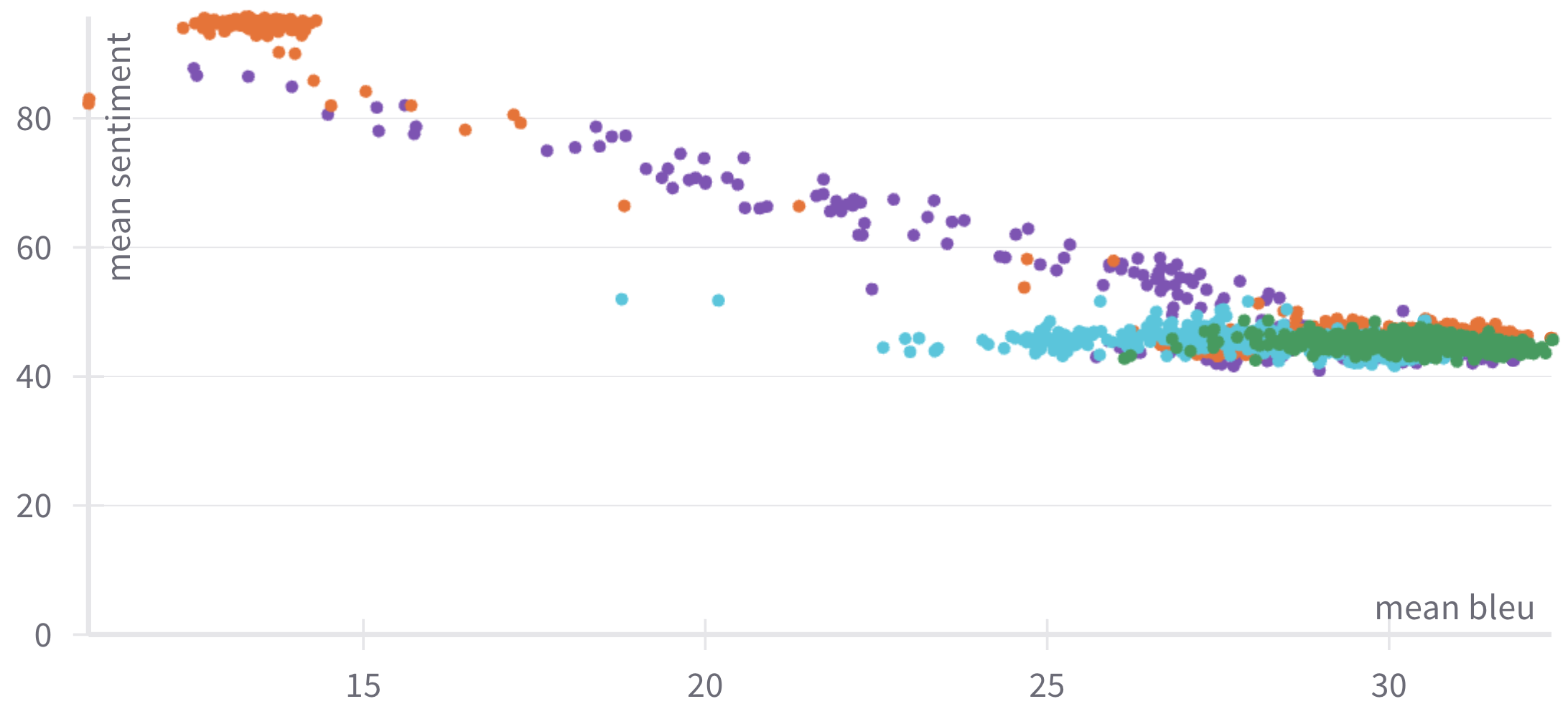}
  \caption*{BLEU vs Sentiment}\label{fig:mt_bleu_sentiment}
\endminipage
\caption{Pairwise reward values for Machine Translation Task from German to English, in different settings of \textcolor{myorange}{average reward}, \textcolor{myblue}{hyper volume indicator reward}, \textcolor{mygreen}{expected product reward}, and \textcolor{mypurple}{multiple gradient descent algorithm}.}  
\label{fig:mt-pairwise}
\end{figure*}

\subsection{Compared Methods}

We compare two volume maximization approaches such as Hyper-volume indicator (HVI) and Expected product of rewards (Product).
We also analyze the Multiple Gradient Descent Algorithm (MGDA) that finds the optimal gradient update direction to maximize all rewards simultaneously.

As a baseline, we use RLPrompt~\cite{deng-etal-2022-rlprompt} for three objectives by optimizing the average of all rewards (Average) and comparing with the above-mentioned approaches.

\begin{table*}[t]
\centering
\scalebox{1.0}{
\begin{tabular}{p{1.5cm} p{3.7cm} p{2.4cm} p{2.4cm} p{0.9cm} p{0.9cm} p{0.9cm} }
\toprule
\textbf{Method} & \textbf{Prompt} &  \textbf{Input} & \textbf{Output} & \textbf{Obj 1} & \textbf{Obj 2} & \textbf{Obj 3} \\
\midrule
\multicolumn{7}{l}{\textcolor{gray}{Text Style Transfer ($Obj_1$: Content - $Obj_2$: Style - $Obj_3$: Sentiment)}} \\
\midrule
HVI & \raggedright thou WhereasYe WhereasOY & \raggedright Be brave like the lion and proud. & \raggedright Be thou like the lion and proud. & 77.4 & 98.14 & 80.06\\
&&&&& \\
& \raggedright EthiopJVphabet wherein whereby  & \raggedright He has to pay the price for his foolish actions. & \raggedright He hath the price for his folly, & 65.2 & 98.48 & 50.26\\
\midrule
MGDA & \raggedright Whereas ye untoWhereasð & \raggedright Don't think I'm in love with him just because I'm asking about him. & \raggedright Thou shalt not think that I'm in love with him just because I'm asking about him. & 81.5 & 89.81 & 51.96 \\
&&&&& \\
& \raggedright SyriJV SHALL unto Ye & \raggedright Take your penalty. & \raggedright Take thy penalty. & 98.03 & 44.14 & 83.3 \\
\midrule
Product & \raggedright whereinWhereasWhereas whereinYe & \raggedright He has to pay the price for his foolish actions. & \raggedright He hath to pay the price for his folly; & 81.1 & 98.36 & 47.07 \\
&&&&& \\
& \raggedright SHALL SHALL hath hath SHALL & \raggedright Yes, my lord. & \raggedright And shall be, my lord. & 55.9 & 97.46 & 84.5 \\
 \bottomrule
\end{tabular}
}
\caption{Given the prompt learned from the policy model alongside the input to GPT2, the Shakespearian form of the sentence is generated as the output. The objective values corresponding to the output, as well as the method used for training the policy model, are reported.}
\label{tab:qualitative-tst}
\end{table*}
\vspace{-1 pt}
\section{Results}

We report individual objective values as well as their expected product and average in Table~ \ref{tab:comparison}.
When evaluating based on the expected product, the product method demonstrates superior performance compared to the other approaches.
Additionally, we observe a more balanced improvement across all rewards with volume-based methods such as HVI and product, in contrast to Average and MGDA.
For example, in style transfer task, the ``average'' method improves \textit{style} disproportionately higher than other objectives despite achieving the best performance based on the Average metric.

Among the compared methods, we believe that expected product of reward best captures the balanced performance across the rewards, as it mitigates the problems of \textit{objective collapse} and \textit{dominant outlier effect}, described in prior sections.





\subsection{Pairwise Reward Analysis}

We plot the pairwise objective values achieved by each of the optimization methods on the validation set for text-style transfer and  machine translation tasks in Figure~\ref{fig:tst-pairwise} and \ref{fig:mt-pairwise} respectively.
Each data point on the scatter plot represents the average objective value computed from 128 output samples, where each output sample is generated from a prompt sampled from the policy network and an input sentence from the validation dataset.
Figure~\ref{fig:tst-pairwise} illustrates how relying on the average of reward values can result in sacrifice of individual objectives in favor of overall improvement. 
We observe instances where sentiment and style scores are notably low, despite a high content score. 
This phenomenon arises due to the emphasis placed solely on the average of rewards, without consideration for individual objectives.
MGDA performs slightly better than the average reward when balancing the individual objectives. 
However, the HVI and the product of rewards improve all the objectives simultaneously, with greater success. 

Similarly, in the case of the machine translation task in Figure~\ref{fig:mt-pairwise}, we observe \textit{objective collapse} for the average reward setting, while the other three approaches demonstrate a better balance among objectives while enhancing the joint reward. 
Notably, the HVI approach and the expected product of rewards are more successful in simultaneously optimizing all the objectives.

\section{Qualitative Analysis}
In this section, we perform a qualitative analysis of the learned prompts.
In Table \ref{tab:qualitative-tst}, we provide two examples of the style transfer task for each method, the generated prompt using it, the input, and the resulting output produced by the frozen model.
The examples in Table \ref{tab:qualitative-tst}, are some of the successful examples chosen based on their high objective scores. In these examples, content is preserved reasonably well, while some of the words are changed in order to be more aligned with the Shakespearean style. The scores corresponding to the positive sentiment scores should be interpreted carefully, as achieving a more positive sentiment might change the semantic meaning of the sentence to some extent, specifically where the original input has an opposite sentiment. For instance, in another example sentence, ``crimes'' was replaced with ``good deeds'' to make the sentiment more positive, at the expense of getting a lower score on the competing objective of content preservation. Overall, the models seem to achieve a balanced performance when handling these conflicting situations. 

Moreover, we can observe some similarities and common words in high-performing prompts, demonstrating the effectiveness of certain tokens for a specific task. However, we can see that despite these similarities, the differences in the prompts from various methods can substantially affect the final evaluation results, which were shown in Table \ref{tab:comparison}.

It is important to note that the highest-performing prompts do not necessarily need to be interpretable by humans. There is value in a discrete prompt beyond it being interpretable by humans: a discrete prompt, unlike a continuous prompt, can be passed into a black-box model like ChatGPT, while a continuous prompt cannot. Furthermore, discrete prompts are more likely to generalize across different LLMs than continuous prompts due to the common text space instead of the model-specific latent space \cite{deng-etal-2022-rlprompt}. In applications where having an interpretable prompt is useful, corresponding rewards could be included in the multi-objective formulation.

\section{Related Work}

\paragraph{Prompt Tuning.}
A line of research has emerged with a focus on improving the discrete~\cite{jiang-etal-2020-know, prasad-etal-2023-grips, mishra-etal-2022-reframing} and soft prompts~\cite{li-liang-2021-prefix, qin-eisner-2021-learning, vu-etal-2022-spot, liu2023gpt} for improved downstream performance.
Few recent works generate discrete prompts by utilizing the models gradients~\cite{shin-etal-2020-autoprompt, wen2023hard}, employing evolution algorithms~\cite{guo2023connecting}, and reinforcement learning~\cite{zhang2023tempera, deng-etal-2022-rlprompt, jung2023discrete, wang2023promptagent}.
Our work shares a similar direction, but we focus on multiple competing objectives instead of one.

\paragraph{Multi-objective Reinforcement Learning.}
Multi-objective reinforcement learning is typically studied in decision-making~\cite{van2013scalarized, van2014multi, yang2019generalized, xu2020prediction, hayes2022practical}.
\citet{jang2023personalized} fine-tunes LMs for multiple objectives by training one policy model per objective and merging them. ~\cite{lin2019pareto, sener2019multitask} perform multi-objective RL in a multi-task learning setup.
Instead, we propose optimizing the prompts for one model with multiple objectives.
\section{Conclusion}
We empirically investigate the use of optimization techniques alongside reinforcement learning to address discrete prompt optimization in a multi-objective context. 
Our experiments show that multi-objective methods, which directly optimize the volume, outperform those seeking monotonic update directions, achieving a better balance across all rewards.

\section{Limitations}
The methods discussed in this paper take many GPU hours to converge, making it computationally expensive to run. 
Moreover, our optimization methods perform well on smaller LMs like GPT2, we have not experimented with larger models because of the substantial computational cost.

\section{Ethical Considerations}
This paper introduces three approaches for discrete prompt optimization. 
As such, prompt-tuning should not introduce biases not already observed in the model and generate any harmful text as prompts, and we do not anticipate any significant ethical concerns.

\clearpage\newpage

\bibliography{anthology,custom}

\begin{thebibliography}{41}
\expandafter\ifx\csname natexlab\endcsname\relax\def\natexlab#1{#1}\fi

\bibitem[{Cettolo et~al.(2017)Cettolo, Federico, Bentivogli, Niehues, St{\"u}ker, Sudoh, Yoshino, and Federmann}]{cettolo2017overview}
Mauro Cettolo, Marcello Federico, Luisa Bentivogli, Jan Niehues, Sebastian St{\"u}ker, Katsuhito Sudoh, Koichiro Yoshino, and Christian Federmann. 2017.
\newblock \href {https://aclanthology.org/2017.iwslt-1.1} {Overview of the {IWSLT} 2017 evaluation campaign}.
\newblock In \emph{Proceedings of the 14th International Conference on Spoken Language Translation}, pages 2--14, Tokyo, Japan. International Workshop on Spoken Language Translation.

\bibitem[{Chung et~al.(2022)Chung, Hou, Longpre, Zoph, Tay, Fedus, Li, Wang, Dehghani, Brahma, Webson, Gu, Dai, Suzgun, Chen, Chowdhery, Narang, Mishra, Yu, Zhao, Huang, Dai, Yu, Petrov, Chi, Dean, Devlin, Roberts, Zhou, Le, and Wei}]{flant5small}
Hyung~Won Chung, Le~Hou, Shayne Longpre, Barret Zoph, Yi~Tay, William Fedus, Eric Li, Xuezhi Wang, Mostafa Dehghani, Siddhartha Brahma, Albert Webson, Shixiang~Shane Gu, Zhuyun Dai, Mirac Suzgun, Xinyun Chen, Aakanksha Chowdhery, Sharan Narang, Gaurav Mishra, Adams Yu, Vincent Zhao, Yanping Huang, Andrew Dai, Hongkun Yu, Slav Petrov, Ed~H. Chi, Jeff Dean, Jacob Devlin, Adam Roberts, Denny Zhou, Quoc~V. Le, and Jason Wei. 2022.
\newblock \href {https://doi.org/10.48550/ARXIV.2210.11416} {Scaling instruction-finetuned language models}.

\bibitem[{Deng et~al.(2022)Deng, Wang, Hsieh, Wang, Guo, Shu, Song, Xing, and Hu}]{deng-etal-2022-rlprompt}
Mingkai Deng, Jianyu Wang, Cheng-Ping Hsieh, Yihan Wang, Han Guo, Tianmin Shu, Meng Song, Eric Xing, and Zhiting Hu. 2022.
\newblock \href {https://doi.org/10.18653/v1/2022.emnlp-main.222} {{RLP}rompt: Optimizing discrete text prompts with reinforcement learning}.
\newblock In \emph{Proceedings of the 2022 Conference on Empirical Methods in Natural Language Processing}, pages 3369--3391, Abu Dhabi, United Arab Emirates. Association for Computational Linguistics.

\bibitem[{Fliege and Svaiter(2000)}]{fliege2000steepest}
J{\"o}rg Fliege and Benar~Fux Svaiter. 2000.
\newblock Steepest descent methods for multicriteria optimization.
\newblock \emph{Mathematical methods of operations research}, 51:479--494.

\bibitem[{Guo et~al.(2022)Guo, Tan, Liu, Xing, and Hu}]{guo2022efficient}
Han Guo, Bowen Tan, Zhengzhong Liu, Eric~P. Xing, and Zhiting Hu. 2022.
\newblock \href {http://arxiv.org/abs/2106.07704} {Efficient (soft) q-learning for text generation with limited good data}.

\bibitem[{Guo et~al.(2023)Guo, Wang, Guo, Li, Song, Tan, Liu, Bian, and Yang}]{guo2023connecting}
Qingyan Guo, Rui Wang, Junliang Guo, Bei Li, Kaitao Song, Xu~Tan, Guoqing Liu, Jiang Bian, and Yujiu Yang. 2023.
\newblock Connecting large language models with evolutionary algorithms yields powerful prompt optimizers.
\newblock \emph{arXiv preprint arXiv:2309.08532}.

\bibitem[{Haddow et~al.(2022)Haddow, Bawden, Miceli~Barone, Helcl, and Birch}]{haddow2022survey}
Barry Haddow, Rachel Bawden, Antonio~Valerio Miceli~Barone, Jind{\v{r}}ich Helcl, and Alexandra Birch. 2022.
\newblock \href {https://doi.org/10.1162/coli_a_00446} {Survey of low-resource machine translation}.
\newblock \emph{Computational Linguistics}, 48(3):673--732.

\bibitem[{Hayes et~al.(2022)Hayes, R{\u{a}}dulescu, Bargiacchi, K{\"a}llstr{\"o}m, Macfarlane, Reymond, Verstraeten, Zintgraf, Dazeley, Heintz et~al.}]{hayes2022practical}
Conor~F Hayes, Roxana R{\u{a}}dulescu, Eugenio Bargiacchi, Johan K{\"a}llstr{\"o}m, Matthew Macfarlane, Mathieu Reymond, Timothy Verstraeten, Luisa~M Zintgraf, Richard Dazeley, Fredrik Heintz, et~al. 2022.
\newblock A practical guide to multi-objective reinforcement learning and planning.
\newblock \emph{Autonomous Agents and Multi-Agent Systems}, 36(1):26.

\bibitem[{He et~al.(2020)He, Wang, Neubig, and Berg-Kirkpatrick}]{he2020APF}
Junxian He, Xinyi Wang, Graham Neubig, and Taylor Berg-Kirkpatrick. 2020.
\newblock \href {https://api.semanticscholar.org/CorpusID:211069439} {A probabilistic formulation of unsupervised text style transfer}.
\newblock \emph{ArXiv}, abs/2002.03912.

\bibitem[{Jang et~al.(2023)Jang, Kim, Lin, Wang, Hessel, Zettlemoyer, Hajishirzi, Choi, and Ammanabrolu}]{jang2023personalized}
Joel Jang, Seungone Kim, Bill~Yuchen Lin, Yizhong Wang, Jack Hessel, Luke Zettlemoyer, Hannaneh Hajishirzi, Yejin Choi, and Prithviraj Ammanabrolu. 2023.
\newblock Personalized soups: Personalized large language model alignment via post-hoc parameter merging.
\newblock \emph{arXiv preprint arXiv:2310.11564}.

\bibitem[{Jhamtani et~al.(2017)Jhamtani, Gangal, Hovy, and Nyberg}]{jhamtani2017shakespearizing}
Harsh Jhamtani, Varun Gangal, Eduard Hovy, and Eric Nyberg. 2017.
\newblock \href {https://doi.org/10.18653/v1/W17-4902} {Shakespearizing modern language using copy-enriched sequence to sequence models}.
\newblock In \emph{Proceedings of the Workshop on Stylistic Variation}, pages 10--19, Copenhagen, Denmark. Association for Computational Linguistics.

\bibitem[{Jiang et~al.(2020)Jiang, Xu, Araki, and Neubig}]{jiang-etal-2020-know}
Zhengbao Jiang, Frank~F. Xu, Jun Araki, and Graham Neubig. 2020.
\newblock \href {https://doi.org/10.1162/tacl_a_00324} {How can we know what language models know?}
\newblock \emph{Transactions of the Association for Computational Linguistics}, 8:423--438.

\bibitem[{Jin et~al.(2022)Jin, Jin, Hu, Vechtomova, and Mihalcea}]{jin2022deep}
Di~Jin, Zhijing Jin, Zhiting Hu, Olga Vechtomova, and Rada Mihalcea. 2022.
\newblock \href {https://doi.org/10.1162/coli_a_00426} {Deep learning for text style transfer: A survey}.
\newblock \emph{Computational Linguistics}, 48(1):155--205.

\bibitem[{Jung and Kim(2023)}]{jung2023discrete}
Hoyoun Jung and Kyung-Joong Kim. 2023.
\newblock Discrete prompt compression with reinforcement learning.
\newblock \emph{arXiv preprint arXiv:2308.08758}.

\bibitem[{Knowles et~al.(2004)Knowles, Corne, and Fleischer}]{knowles2004bounded}
Joshua Knowles, David Corne, and Mark Fleischer. 2004.
\newblock \href {https://doi.org/10.1109/CEC.2003.1299401} {Bounded archiving using the lebesgue measure}.

\bibitem[{Li and Liang(2021)}]{li-liang-2021-prefix}
Xiang~Lisa Li and Percy Liang. 2021.
\newblock \href {https://doi.org/10.18653/v1/2021.acl-long.353} {Prefix-tuning: Optimizing continuous prompts for generation}.
\newblock In \emph{Proceedings of the 59th Annual Meeting of the Association for Computational Linguistics and the 11th International Joint Conference on Natural Language Processing (Volume 1: Long Papers)}, pages 4582--4597, Online. Association for Computational Linguistics.

\bibitem[{Lin et~al.(2019)Lin, Zhen, Li, Zhang, and Kwong}]{lin2019pareto}
Xi~Lin, Hui-Ling Zhen, Zhenhua Li, Qing-Fu Zhang, and Sam Kwong. 2019.
\newblock \href {https://proceedings.neurips.cc/paper_files/paper/2019/file/685bfde03eb646c27ed565881917c71c-Paper.pdf} {Pareto multi-task learning}.
\newblock In \emph{Advances in Neural Information Processing Systems}, volume~32. Curran Associates, Inc.

\bibitem[{Liu et~al.(2023)Liu, Zheng, Du, Ding, Qian, Yang, and Tang}]{liu2023gpt}
Xiao Liu, Yanan Zheng, Zhengxiao Du, Ming Ding, Yujie Qian, Zhilin Yang, and Jie Tang. 2023.
\newblock Gpt understands, too.
\newblock \emph{AI Open}.

\bibitem[{Mishra et~al.(2022)Mishra, Khashabi, Baral, Choi, and Hajishirzi}]{mishra-etal-2022-reframing}
Swaroop Mishra, Daniel Khashabi, Chitta Baral, Yejin Choi, and Hannaneh Hajishirzi. 2022.
\newblock \href {https://doi.org/10.18653/v1/2022.findings-acl.50} {Reframing instructional prompts to {GPT}k{'}s language}.
\newblock In \emph{Findings of the Association for Computational Linguistics: ACL 2022}, pages 589--612, Dublin, Ireland. Association for Computational Linguistics.

\bibitem[{Papineni et~al.(2002)Papineni, Roukos, Ward, and jing Zhu}]{papineni02bleu}
Kishore Papineni, Salim Roukos, Todd Ward, and Wei jing Zhu. 2002.
\newblock Bleu: a method for automatic evaluation of machine translation.

\bibitem[{Prasad et~al.(2023)Prasad, Hase, Zhou, and Bansal}]{prasad-etal-2023-grips}
Archiki Prasad, Peter Hase, Xiang Zhou, and Mohit Bansal. 2023.
\newblock \href {https://doi.org/10.18653/v1/2023.eacl-main.277} {{G}r{IPS}: Gradient-free, edit-based instruction search for prompting large language models}.
\newblock In \emph{Proceedings of the 17th Conference of the European Chapter of the Association for Computational Linguistics}, pages 3845--3864, Dubrovnik, Croatia. Association for Computational Linguistics.

\bibitem[{Qin and Eisner(2021)}]{qin-eisner-2021-learning}
Guanghui Qin and Jason Eisner. 2021.
\newblock \href {https://doi.org/10.18653/v1/2021.naacl-main.410} {Learning how to ask: Querying {LM}s with mixtures of soft prompts}.
\newblock In \emph{Proceedings of the 2021 Conference of the North American Chapter of the Association for Computational Linguistics: Human Language Technologies}, pages 5203--5212, Online. Association for Computational Linguistics.

\bibitem[{Radford et~al.(2019)Radford, Wu, Child, Luan, Amodei, and Sutskever}]{Radford2019LanguageMA}
Alec Radford, Jeff Wu, Rewon Child, David Luan, Dario Amodei, and Ilya Sutskever. 2019.
\newblock Language models are unsupervised multitask learners.

\bibitem[{Sanh et~al.(2019)Sanh, Debut, Chaumond, and Wolf}]{sanh2019distilbert}
Victor Sanh, Lysandre Debut, Julien Chaumond, and Thomas Wolf. 2019.
\newblock Distilbert, a distilled version of bert: smaller, faster, cheaper and lighter.
\newblock In \emph{NeurIPS EMC2 Workshop}.

\bibitem[{Schick and Sch{\"u}tze(2020)}]{Schick2020ExploitingCF}
Timo Schick and Hinrich Sch{\"u}tze. 2020.
\newblock \href {https://api.semanticscholar.org/CorpusID:210838924} {Exploiting cloze-questions for few-shot text classification and natural language inference}.
\newblock In \emph{Conference of the European Chapter of the Association for Computational Linguistics}.

\bibitem[{Sener and Koltun(2019)}]{sener2019multitask}
Ozan Sener and Vladlen Koltun. 2019.
\newblock \href {http://arxiv.org/abs/1810.04650} {Multi-task learning as multi-objective optimization}.

\bibitem[{Shakespeare(1703)}]{shakespeare1703tragedy}
William Shakespeare. 1703.
\newblock \emph{The tragedy of Hamlet, prince of Denmark}.
\newblock Wellington.

\bibitem[{Shakespeare(1710)}]{shakespeare1710macbeth}
William Shakespeare. 1710.
\newblock \emph{Macbeth. A tragedy. With all the alterations, amendments, additions, and new songs [by Sir William Davenant], as it is now acted at the Queen's Theatre.[Anonymous.]}.
\newblock J. Tonson.

\bibitem[{Shakespeare(1734)}]{shakespeare1734midsummer}
William Shakespeare. 1734.
\newblock \emph{A Midsummer Night's Dream}, volume~2.
\newblock J. TONSON, and the rest of the PROPRIETORS; and sold.

\bibitem[{Shin et~al.(2020)Shin, Razeghi, Logan~IV, Wallace, and Singh}]{shin-etal-2020-autoprompt}
Taylor Shin, Yasaman Razeghi, Robert~L. Logan~IV, Eric Wallace, and Sameer Singh. 2020.
\newblock \href {https://doi.org/10.18653/v1/2020.emnlp-main.346} {{A}uto{P}rompt: {E}liciting {K}nowledge from {L}anguage {M}odels with {A}utomatically {G}enerated {P}rompts}.
\newblock In \emph{Proceedings of the 2020 Conference on Empirical Methods in Natural Language Processing (EMNLP)}, pages 4222--4235, Online. Association for Computational Linguistics.

\bibitem[{Van~Moffaert et~al.(2013)Van~Moffaert, Drugan, and Now{\'e}}]{van2013scalarized}
Kristof Van~Moffaert, Madalina~M Drugan, and Ann Now{\'e}. 2013.
\newblock Scalarized multi-objective reinforcement learning: Novel design techniques.
\newblock In \emph{2013 IEEE symposium on adaptive dynamic programming and reinforcement learning (ADPRL)}, pages 191--199. IEEE.

\bibitem[{Van~Moffaert and Now{\'e}(2014)}]{van2014multi}
Kristof Van~Moffaert and Ann Now{\'e}. 2014.
\newblock Multi-objective reinforcement learning using sets of pareto dominating policies.
\newblock \emph{The Journal of Machine Learning Research}, 15(1):3483--3512.

\bibitem[{Vu et~al.(2022)Vu, Lester, Constant, Al-Rfou{'}, and Cer}]{vu-etal-2022-spot}
Tu~Vu, Brian Lester, Noah Constant, Rami Al-Rfou{'}, and Daniel Cer. 2022.
\newblock \href {https://doi.org/10.18653/v1/2022.acl-long.346} {{SP}o{T}: Better frozen model adaptation through soft prompt transfer}.
\newblock In \emph{Proceedings of the 60th Annual Meeting of the Association for Computational Linguistics (Volume 1: Long Papers)}, pages 5039--5059, Dublin, Ireland. Association for Computational Linguistics.

\bibitem[{Wang et~al.(2023)Wang, Li, Wang, Bai, Luo, Zhang, Jojic, Xing, and Hu}]{wang2023promptagent}
Xinyuan Wang, Chenxi Li, Zhen Wang, Fan Bai, Haotian Luo, Jiayou Zhang, Nebojsa Jojic, Eric~P Xing, and Zhiting Hu. 2023.
\newblock Promptagent: Strategic planning with language models enables expert-level prompt optimization.
\newblock \emph{arXiv preprint arXiv:2310.16427}.

\bibitem[{Wen et~al.(2023)Wen, Jain, Kirchenbauer, Goldblum, Geiping, and Goldstein}]{wen2023hard}
Yuxin Wen, Neel Jain, John Kirchenbauer, Micah Goldblum, Jonas Geiping, and Tom Goldstein. 2023.
\newblock Hard prompts made easy: Gradient-based discrete optimization for prompt tuning and discovery.
\newblock \emph{arXiv preprint arXiv:2302.03668}.

\bibitem[{Xu et~al.(2020)Xu, Tian, Ma, Rus, Sueda, and Matusik}]{xu2020prediction}
Jie Xu, Yunsheng Tian, Pingchuan Ma, Daniela Rus, Shinjiro Sueda, and Wojciech Matusik. 2020.
\newblock Prediction-guided multi-objective reinforcement learning for continuous robot control.
\newblock In \emph{International conference on machine learning}, pages 10607--10616. PMLR.

\bibitem[{Xu et~al.(2012)Xu, Ritter, Dolan, Grishman, and Cherry}]{xu2012paraphrasing}
Wei Xu, Alan Ritter, Bill Dolan, Ralph Grishman, and Colin Cherry. 2012.
\newblock \href {https://aclanthology.org/C12-1177} {Paraphrasing for style}.
\newblock In \emph{Proceedings of {COLING} 2012}, pages 2899--2914, Mumbai, India. The COLING 2012 Organizing Committee.

\bibitem[{Yang et~al.(2019)Yang, Sun, and Narasimhan}]{yang2019generalized}
Runzhe Yang, Xingyuan Sun, and Karthik Narasimhan. 2019.
\newblock A generalized algorithm for multi-objective reinforcement learning and policy adaptation.
\newblock \emph{Advances in neural information processing systems}, 32.

\bibitem[{Zhang et~al.(2023)Zhang, Wang, Zhou, Schuurmans, and Gonzalez}]{zhang2023tempera}
Tianjun Zhang, Xuezhi Wang, Denny Zhou, Dale Schuurmans, and Joseph~E. Gonzalez. 2023.
\newblock \href {https://openreview.net/forum?id=gSHyqBijPFO} {{TEMPERA}: Test-time prompt editing via reinforcement learning}.
\newblock In \emph{The Eleventh International Conference on Learning Representations}.

\bibitem[{Zhang et~al.(2020)Zhang, Kishore, Wu, Weinberger, and Artzi}]{zhang2020bertscore}
Tianyi Zhang, Varsha Kishore, Felix Wu, Kilian~Q. Weinberger, and Yoav Artzi. 2020.
\newblock \href {http://arxiv.org/abs/1904.09675} {Bertscore: Evaluating text generation with bert}.

\bibitem[{Zitzler and Thiele(1998)}]{zitzler1998multiobj}
Eckart Zitzler and Lothar Thiele. 1998.
\newblock Multiobjective optimization using evolutionary algorithms --- a comparative case study.
\newblock In \emph{Parallel Problem Solving from Nature --- PPSN V}, pages 292--301, Berlin, Heidelberg. Springer Berlin Heidelberg.

\end{thebibliography}

\newpage
\appendix

\clearpage\newpage
\section{Appendix}

\subsection{Multiple Gradient Descent Algorithm}
\label{app:mgda}

~\cite{fliege2000steepest} proposes a steepest descent algorithm for multi-criteria optimization, where the update rule for the parameters $\theta$ at time $t$ with the step size $\eta$ is defined as:
\begin{equation}
    \theta_{t+1}=\theta_{t}-\eta d_{t}
\end{equation}
where the search direction $d_{t}$ is calculated as follows, with $\mathcal{L}_i(\theta_j)$ being the expected loss corresponding to objective $o_i$:
\begin{equation}
\label{eqn:steep-gradient-descent-app}
\begin{aligned}
    &\left(d_{t}, \alpha_{t}\right)=\arg \min _{d \in \mathbb{R}^n, \alpha \in \mathbb{R}} \alpha+\frac{1}{2}\|d\|^2, \\
    &\text{s.t. } \nabla \mathcal{L}_i\left(\theta_t\right)^T d \leq \alpha, \quad i=1, \ldots, m .
\end{aligned}
\end{equation}

A valid direction $d_{t}$ improves the values for all the objectives, simultaneously. 
Moreover, ~\cite{fliege2000steepest} shows that the solution obtained by the aforementioned approach leads to a Pareto critical point.

Based on the KKT conditions, we have
\begin{equation}
d_t=-\left(\sum_{i=1}^m \lambda_i \nabla \mathcal{L}_i\left(\theta_t\right)\right), \quad \sum_{i=1}^m \lambda_i=1
\end{equation}
and we can write equation-\ref{eqn:steep-gradient-descent-app} in its dual form:

\begin{equation}
\label{eqn:dual}
\begin{aligned}
\max _{\lambda_i} & -\frac{1}{2}\left\|\sum_{i=1}^m \lambda_i \nabla \mathcal{L}_i\left(\theta_t\right)\right\|^2 \\
\text { s.t. } & \sum_{i=1}^m \lambda_i=1, \lambda_i \geq 0,, \forall i=1, \ldots, m.
\end{aligned}
\end{equation}

Therefore, in order to find a valid direction d that improves all the objectives, we can rewrite the equation in its dual form. This will give us a constrained optimization problem, which can be solved by the Frank-Wolfe algorithm, and we can then use gradient descent to update the policy parameters.

\end{document}